\documentclass[runningheads]{llncs}
\usepackage{graphicx}
\usepackage[hidelinks]{hyperref}
\usepackage{amsmath}
\usepackage{float}
\usepackage{amsfonts}
\usepackage[caption=false]{subfig}
\usepackage{multirow}
\usepackage{todonotes}
\usepackage{xcolor}
\usepackage{multirow}
\presetkeys%
{todonotes}%
{inline,backgroundcolor=yellow}{}
\usepackage{xargs}                      

\newcommandx{\important}[2][1=]{\todo[linecolor=red,backgroundcolor=red!25,bordercolor=red,#1]{\textbf{Important:}#2}}

\newcommandx{\eirini}[2][1=]{\todo[linecolor=green,backgroundcolor=green!25,bordercolor=green,#1]{\textbf{Eirini:}#2}}
\newcommandx{\manolis}[2][1=]
{\todo[linecolor=gray,backgroundcolor=gray!25,bordercolor=gray,#1]{\textbf{Manolis:}#2}}
\newcommandx{\arjun}[2][1=]
{\todo[linecolor=blue,backgroundcolor=blue!25,bordercolor=blue,#1]{\textbf{Arjun:}#2}}

\begin{document}
\title{Learning impartial policies for sequential counterfactual explanations using Deep Reinforcement Learning}
\titlerunning{Sequential counterfactual policies using RL}
%

\author{}
\institute{}

\author{Emmanouil Panagiotou\inst{1,2}\orcidID{0000-0001-9134-9387} \and
Eirini Ntoutsi\inst{2}\orcidID{0000-0001-5729-1003}}
\authorrunning{E. Panagiotou et al.}
%

\institute{Freie Universität Berlin, Department of Mathematics \& Computer Science, Berlin, Germany \\
\email{emmanouil.panagiotou@fu-berlin.de}\\
\and
Universität der Bundeswehr München, Faculty for Informatik,  Munich, Germany
}

\maketitle              
\begin{abstract}

In the field of explainable Artificial Intelligence (XAI), sequential counterfactual (SCF) examples are often used to alter the decision of a trained classifier by implementing a sequence of modifications to the input instance.
Although certain test-time algorithms aim to optimize for each new instance individually, recently Reinforcement Learning (RL) methods have been proposed that seek to learn policies for discovering SCFs, thereby enhancing scalability.
As is typical in RL, the formulation of the RL problem, including the specification of state space, actions, and rewards, can often be ambiguous.
In this work, we identify shortcomings in existing methods that can result in policies with undesired properties, such as a bias towards specific actions. We propose to use the output probabilities of the classifier to create a more informative reward, to mitigate this effect. 

\end{abstract}
\section{Introduction}
\label{sec:introduction}
Predictive Machine Learning (ML) models have been used in various fields to make predictions by generalizing knowledge learned from data. Yet, in recent years, generating explanations for model decisions has become increasingly important, shifting interest to the field of explainable AI (XAI) \cite{guidotti2018survey}. A popular category of XAI methods are the so-called ``counterfactual explanations'' that offer a "what-if" analysis of model decisions by identifying the minimal set of changes or interventions needed to alter the outcome of the model for a given instance, allowing users to understand the factors that influenced the model's decision.
Common desiderata for counterfactual explanations, as highlighted by \cite{dandl2020multi}, include \emph{proximity} (distance) to the original instance, \emph{sparsity} in terms of changes made, and the \emph{plausibility} of the counterfactual instances w.r.t. the particular problem domain.

Recently, a specific type of counterfactual explanations, called sequential counterfactuals (SCFs) has been proposed~\cite{CF_consequence,CF_sequences,CF_sequential_uncertainty,CF_constraints,CF_tabular,CF_gower} that, rather than making instantaneous or simultaneous modifications to the input, propose a sequence of alterations, each building upon the previous one, until the classifier's decision changes. Such approaches allow for considering the consequences of certain changes to other attributes, for instance, increasing the age of an individual might require an increase in their level of education to maintain consistency or plausibility.
SCF generation algorithms look for the optimal order of changes to an input instance, to alter the decision of a model.

In the generation of SCFs, certain methods employ search algorithms that need to initiate optimization for each new instance \cite{CF_consequence,CF_case_based}, leading to efficiency problems. On the other hand, feed-forward methods~\cite{amortized_discrete_actions,relax_continuous_actions} leverage Reinforcement Learning (RL) to learn a scalable policy for generating SCFs. 
The problem is formulated as a Markov Decision Process (MDP), where \emph{states} present the original input instance and its alternations, and \emph{actions} correspond to all allowed changes to input features (discrete or continuous). The immediate \emph{reward} for taking an action depends on the output of the black-box model and on the distance of the altered instance to its original state (proximity), to ensure minimal feature changes (action sparsity). 

As is typical in RL, the problem formulation plays a crucial role in shaping the policy that the agent learns. In this study, we focus on analyzing certain modeling choices, specifically those related to the action space and the shaping of the reward function, which is influenced by the output of the classifier. We observe that rewarding the RL agent based solely on the sparse binary decision of the classifier can result in problematic policies that only modify features highly correlated with the classifier's output.
While these methods demonstrate satisfactory performance in terms of action sparsity, satisfiability, and distance, they prove to be ineffective in practice as they learn to take identical actions for any given input, leading to what we call \emph{feature over-utilization}. In extreme cases, this phenomenon leads to altering a single feature for any given input, thereby reducing the practicality and diversity of generated counterfactual explanations. We present an action entropy metric as a means to quantify the extent of feature over-utilization. Moreover, we propose to overcome this issue by taking continuous actions and implementing a denser reward function when the classifier's output probability is available instead of solely relying on the class labels.

The rest of this paper is organized as follows: related work is discussed in Section \ref{sec:related}. In Section \ref{sec:framework} we introduce our proposed method. Experimental results are provided in Section \ref{sec:experiments}. Conclusions and future work are discussed in Section \ref{sec:conclusion}.

\section{Related Work}
\label{sec:related}
In recent years, many XAI techniques have been proposed to solve the problem of finding SCFs for a pre-trained model. Model-agnostic methods are more generalizable, as they do not require differentiability or information on the internal architecture of the models, but only use the output for decision-making. The majority are search algorithms that re-optimize for each new given instance, for example by searching in the neighborhood \cite{yi,nice}, or by population-based approaches \cite{CF_consequence,mocf}. These algorithms start from a given instance and navigate through trial and error to find a CF, often optimizing for multiple objectives. In our setting, a RL policy has to be learned that can be applied as a feed-forward method after training, on any given instance.

In \cite{amortized_discrete_actions} a Deep Q-Network (DQN) is used to predict discrete feature changes, which are defined as constant steps. Although the reward function incorporates classifier probabilities whenever feasible, its effect on the SCF generation is not discussed. Besides, the use of fixed increments presents issues by assuming all discrete features to be ordinal and restricting changes in continuous features to discrete steps only. This is further detailed in Section \ref{sec:framework}.

To overcome this problem, \cite{relax_continuous_actions} proposes to use a parametrized DQN (P-DQN) \cite{PDQN} network, which combines a DQN with a policy gradient method (DDPG) \cite{DDPG}, that can be described as the continuous counterpart of a DQN. Doing that, P-DQN can choose among all features using the DQN, and predict a more precise continuous feature change using the DDPG network. Nonetheless, the reward depends only on the class labels, which leads to over-utilization of some features as presented in our experiments.

\section{Overcoming feature over-utilization through informative reward design}
\label{sec:framework}
In this section, we address the issue of action over-utilization that arises when employing sparse rewards in RL techniques for generating SCFs. To overcome this issue, we propose the adoption of a denser reward function that employs the probabilities obtained from the black-box classifier.

\subsection{Limitations of existing approaches}

\noindent{\textbf{DQN:}}
As in \cite{amortized_discrete_actions}, a DQN takes discrete steps of ($\pm 0.05$) for continuous features that are in the range of $[-1,1]$, and ($\pm 1$) for discrete features. This formulation is problematic because i) some discrete features can be nominal, e.g. Occupations don't necessarily have a specific ordering, ii) for continuous features, the step size is discrete and constant, e.g. working hours $\pm 30 min$,  and iii) the constant step size of $\pm 0.05$ does not take feature interdependencies into account, for example, it can translate to $\pm 30 min$ for working hours, but $\pm 2500 EUR$ for capital gain, depending on the normalization method. The paper mentions that a dense reward using the black-box probabilities is applied when available. Furthermore, results report a very low action sparsity (close to $1$) and distance ($0.04$), which is desired when generating SCFs, as a lesser number of changes, closer to the original instance, are more interpretable. However, this also means that for all input instances, CFs are found after changing a single feature just once, as the distance is almost the step size of $0.05$, leading to feature over-utilization and scarcity of diverse options. Such an unvarying single-step policy does not provide useful SCF explanations, and potentially other XAI approaches such as global feature importance methods \cite{dice,shap} are more fitting in this case. Due to these many disadvantages, we do not follow this approach.   

\noindent{\textbf{P-DQN:}}
As proposed in \cite{relax_continuous_actions} the P-DQN \cite{PDQN} model is used to perform continuous changes sequentially. Taking an action consists of a high-level feature choice $k$ (discrete) and the new value of that feature $u_k$ (continuous). This assumption considers all feature changes $u_k$ to be continuous, which necessitates discretization when handling discrete features. However, P-DQN remains more accurate and adaptable compared to the DQN method, as it enables continuous changes for continuous features without the restriction of discrete steps. Analytically, the action space for P-DQN is defined as:

\begin{equation}
A = \Big\{ (k, u_k) \mid u_k \in [-1,1]  \quad k \in [K] \Big\}
\label{eqn:pdqn_action_space}
\end{equation}
 
Where $[K] = \{1, 2, ... , K\}$ are the different features. Each time a feature is changed using an action, the timestep $t$ is incremented. A state $s_t = (x_t, b_t)$ is defined as a tuple of some instance $x_t$ concatenated with a binary indicator vector of previously used actions $b_t$. When taking an action $a_t=(k, u_k)$ the new state $s_{t+1} = (x_{t+1}, b_{t+1})$ is created by updating the value of the chosen feature $k$ as $u_k$, i.e. $x_{t+1}[k] = u_k$, and incrementing the indicator vector $b_{t+1}[k] \mathrel{{+}{=}} 1$. 
Finally, the reward function used in this method \cite{relax_continuous_actions} is defined as:

\begin{equation}
\label{eqn:reward}
R^{bin}_{t} =
\left\{\!\begin{aligned}
& \textit{Pen} &\text{ if } s_{t+1}= \textnormal{failure},\\[1ex]
& \textit{Pos} &\text{ if } s_{t+1}=  \textnormal{success},\\[1ex]
& 0 & else.
\end{aligned}\right\}
- \beta * \delta(s_{t+1}, s_0)
\end{equation}

Where \textit{Pen} is a negative penalty if a failure terminal state is reached, i.e. violating a feature constraint or re-using the same action, \textit{Pos} is a large positive reward if a success terminal state is reached (counterfactual found), $\delta()$ is a distance measure (the $l_2$ norm in our case), and $\beta$ is a parameter controlling the weight of the distance measure. It is evident that this reward function is sparse since it only has access to the binary class labels and not the probabilities. Therefore we refer to this reward function as $R^{bin}$. In Section \ref{sec:experiments}, we demonstrate that the learned policy using this reward is suboptimal in terms of action variety. 

\subsection{P-DQN \& classifier probability rewards}

To construct a more dense reward, we assume that the classifier $ f(\mathcal{X}) = \mathcal{Y}$ makes a binary decision $\mathcal{Y}$ given an input $\mathcal{X}$ based on a probability measure. We denote the probability of an instance belonging to the target class as $P(\mathcal{X}) = P(f(\mathcal{X}) = \textnormal{target})$. This way we can redefine the reward function for a non-terminal state as follows:
\begin{equation}
\label{eqn:reward_new}
R^{prob}_{t} = \alpha * \Bigl(P(s_{t+1}) - P(s_0)\Bigr) - \beta * \delta(s_{t+1}, s_0)
\end{equation}
Where $P(s_{t+1}) - P(s_0)$ is the difference in target class probability between the next state and the initial state and $\alpha, \beta$ are parameters controlling the balance between the difference of probability and distance respectively. This is beneficial because the RL agent has immediate feedback on each step regarding the direction taken in relation to the target class. In contrast, when using the aforementioned binary reward ($R^{bin}$), the agent must explore the space without receiving any positive feedback until a successful state is reached (found CF). This additional information on every step results in a more dense reward function, that not only discourages large changes but also guides towards the target class. We denote this reward function with $R^{prob}$ since it incorporates the utilization of target probabilities.

\section{Experiments \& Results}
\label{sec:experiments}
In this section, we compare the P-DQN model's performance when using the binary reward $R^{bin}$ and when employing the target class probabilities reward $R^{prob}$.

\subsection{Datasets}
Motivated by previous approaches we use four tabular datasets in our comparative study. Specifically, the Adult Income, Credit Approval, German Credit, and German Risk \cite{uci}. Each dataset has a binary output label and is comprised of mixed features. Some datasets also have immutable features (e.g. Race) that are considered when defining the states but are not taken into account in the action space.

\begin{table}[ht]
\centering
\caption{Characteristics of the four public datasets used.}
\begin{tabular}{|c|c|c|c|c|}
\hline
\textbf{Dataset} & \textbf{N. instances} & \multicolumn{3}{|c|}{\textbf{Features}}\\
\hline
 &  & N. of Continuous & N. of Discrete & N. of Immutable\\ 
\hline
Adult Income & 48842 & 4 & 3 & 4\\ 
Credit Approval & 690 & 5 & 10 & 0\\ 
German Credit & 1000 & 5 & 11 & 4\\
German Risk & 1000 & 3 & 4 & 2\\ 
\hline
\end{tabular}
\label{tab:datasets}
\end{table}

\subsection{Evaluation measures}
Based on previous works on SCF generation \cite{relax_continuous_actions,amortized_discrete_actions,CF_consequence}, we use the metrics of satisfiability, distance (i.e. proximity), and action sparsity to evaluate the quality of the generated CF examples. Additionally, we measure the target probability of the found CF and define a novel entropy metric to evaluate the diversity of the actions taken by the policy. When an agent has learned to modify only a specific feature for any given instance, the entropy metric will be low. Conversely, policies with higher entropy exhibit a tendency to modify a diverse range of features depending on the input provided.
All metrics are analytically defined hereafter:
\begin{itemize}
    \item CF $\textnormal{satisfiability} = \frac{\textnormal{N. of counterfactuals found}}{\textnormal{N. of test instances}}$, measures the percentage of counterfactuals found, i.e. ended in a success state.
    \item Action sparsity = N. of features changed to reach a CF (averaged over all test instances), no repeating actions allowed.
    \item $l_2$ norm distance of the CF to the original state (averaged over all test instances)
    \item entropy  $\mathbb{H} = -\frac{\sum\limits_{t}\sum\limits_{k} p^k_t \log_2(p^k_t)}{\log_2(K^2)}$ where $p^k_t = \frac{\textnormal{N. of times feature } k \textnormal{ was changed at step } t}{\textnormal{N. of total feature changes}}$
\end{itemize}

We calculate the entropy of an agent by assigning \textit{probabilities} $p^k_t$ for every feature $k$ to change at each step $t$. We accomplish this by counting how many times the agent changed a feature at each step and dividing by the total number of changes. We normalize dividing by the maximum possible entropy $H_{max}= \log_2(K^2)$, where $K$ corresponds to the total number of mutable features, which is equal to the maximum timesteps, considering that repeating actions are prohibited.

\subsection{Experimental setup}
Regarding the reward functions of Eq. \ref{eqn:reward}, \ref{eqn:reward_new}, the distance metric used is $\delta = l_2$ norm, and the hyper-parameters $(\alpha, \beta)=(10, 1)$ are set empirically to give more weight to the change of probability, given that distance values can be larger. We set the large positive reward for a found CF as $\textit{Pos}=\alpha*thr = 5$ where $thr=0.5$ is the probability threshold of the classifier, and the large negative penalty for a failure state, $\textit{Pen} = -2*\textit{Pos} = -10$.    
We use $80\%$ of the data for training across $40.000$ environment episodes and $20\%$ for evaluation. Each experiment is repeated five times with different random seeds. Experimental results are reported as the average values and standard deviations over all seed initializations.

\subsection{Results}

We compare the P-DQN model using the binary reward used in \cite{relax_continuous_actions}, denoted as $R^{bin}$, to our proposed reward $R^{prob}$ that we define in Section \ref{sec:framework}. Results are shown in Table \ref{tab:eval}.

It is evident that the original approach (P-DQN $R^{bin}$) is mostly performing better in terms of satisfiability, sparsity, and distance. This indicates that it finds SCFs more consistently (satisfiability) for most test instances while taking fewer steps to achieve that (action sparsity) and with close proximity to the original instance. Nevertheless, the entropy of the policy learned by this method is low for all datasets, i.e. under $63\%$, compared to our approach where the policy has higher entropy over $68\%$. 

It should be emphasized that the objectives of action sparsity and entropy can be contradicting. In particular, a policy that consistently alters a single feature strongly correlated with the output target class may exhibit a desired low sparsity metric but also underperforms with a significantly low entropy value. This occurs for example for the Adult Income dataset where P-DQN $R^{bin}$ performs very well at action sparsity but with a huge loss in entropy.

\begin{figure}[H]
  \centering
  \subfloat[P-DQN $R^{bin}$]{
    \includegraphics[width=\textwidth]{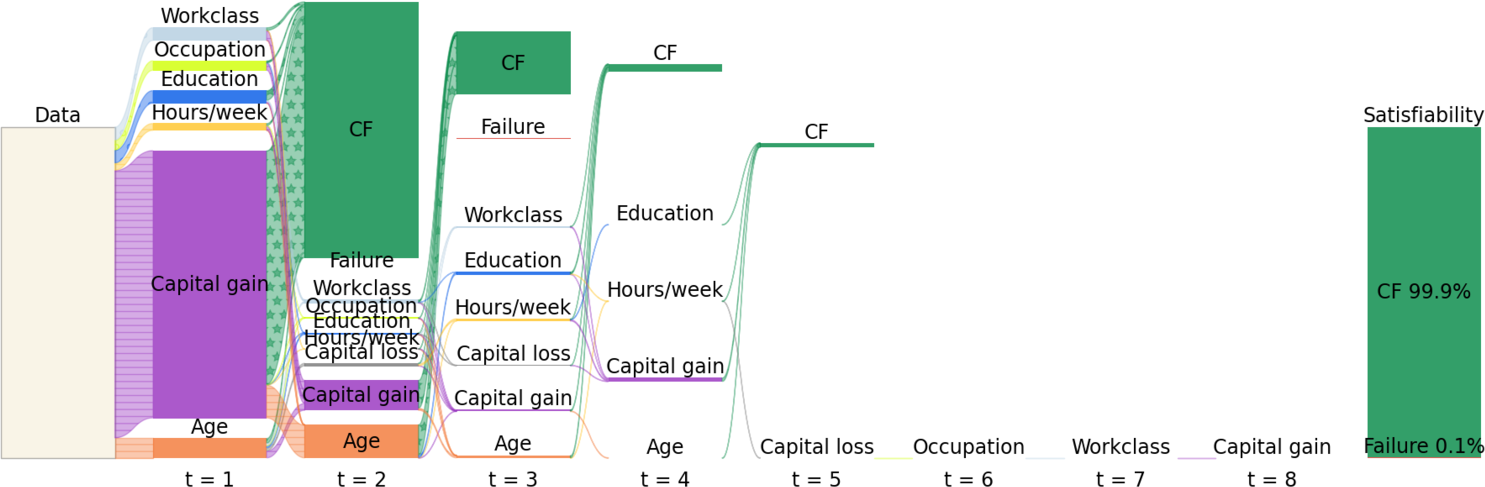}
  }
  \hfill
  \subfloat[P-DQN $R^{prob}$]{
    \includegraphics[width=\textwidth]{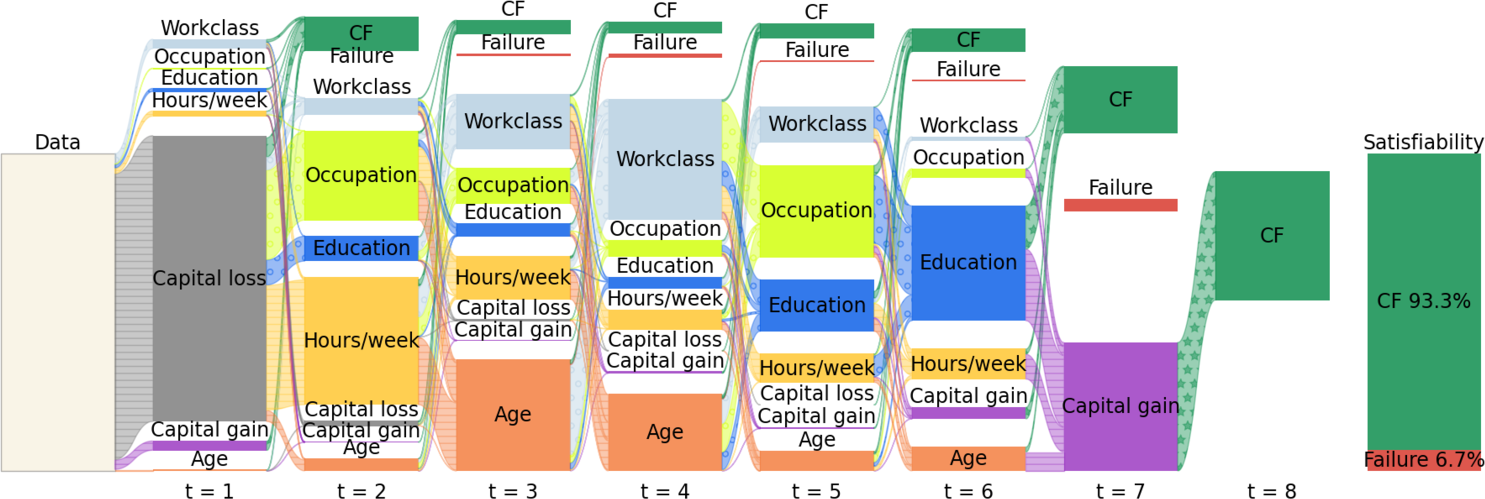}
  }
  \caption{Sankey diagrams for (a) binary reward and (b) our target probability reward.}
  \label{fig:sankey}
\end{figure}

We can further visualize this issue by plotting the Sankey diagrams for both policies in Figure \ref{fig:sankey}.
Sankey diagrams are used to depict the flow of information between sets of values called nodes.

In our case, the x-axis corresponds to the distinct timesteps of the agent during the evaluation episodes. On the y-axis, the first node represents all initial instances of the test data. For each test instance and at each step, the trained RL agent can alter a single feature, causing the flow to split into different nodes. This process continues until a CF is found (green), or a failure state is reached (red). The final node represents the percentage of CFs found, i.e. the satisfiability. We can see that our approach finds SCFs by using a much greater variety of actions, due to the denser reward function that incorporates the target probabilities.


\begin{table}[H]
\centering
\caption{Comparison of both methods on all evaluation metrics and datasets. The results are presented as the average (with standard deviation) over five random initializations for each experiment.}
\label{tab:eval}
\scalebox{0.9}{

\begin{tabular}{|c|c|c|c|c|c|c|c|}
\hline
 \textbf{Dataset} & \textbf{Method} & \textbf{Satisfiability} $\uparrow$ & \textbf{Sparsity} $\downarrow$ & \textbf{distance $\delta$}  $\downarrow$ & \textbf{$P()$} $\uparrow$ & \textbf{entropy $\mathbb{H}$}  $\uparrow$ \\
\hline
\multirow{2}{*}{\shortstack{Adult \\ Income}} & P-DQN $R^{bin}$ & \textbf{1.00} (0.00) & \textbf{1.18} (0.06) & \textbf{1.25} (0.04) & \textbf{0.75} (0.02) & 0.29 (0.06) \\
\cline{2-7}
& P-DQN $R^{prob}$ & 0.92 (0.04) & 5.35 (0.09) & 2.69 (0.23) & 0.66 (0.03) & \textbf{0.82} (0.03)\\
\hline
\multirow{2}{*}{\shortstack{Credit \\ Approval}} & P-DQN $R^{bin}$ & 0.80 (0.40) & \textbf{1.77} (0.15) & \textbf{1.25} (0.65) & 0.55 (0.21) & 0.45 (0.07) \\
\cline{2-7}
& P-DQN $R^{prob}$ & \textbf{0.93} (0.02) & 8.16 (2.04) & 3.94 (0.74) & \textbf{0.65} (0.05) & \textbf{0.84} (0.01) \\
\hline
\multirow{2}{*}{\shortstack{German \\ Credit}} & P-DQN $R^{bin}$ & 0.97 (0.05) & \textbf{1.46} (0.05) & \textbf{1.39} (0.04) & \textbf{0.70} (0.02) & 0.58 (0.02) \\
\cline{2-7}
& P-DQN $R^{prob}$ & \textbf{0.97} (0.01) & 1.71 (0.16) & 1.47 (0.06) & 0.69 (0.02) & \textbf{0.68} (0.05) \\
\hline
\multirow{2}{*}{\shortstack{German \\ Risk}} & P-DQN $R^{bin}$ & \textbf{0.97} (0.03) & \textbf{1.81} (0.37) & \textbf{1.52} (0.11) & \textbf{0.63} (0.02) & 0.67 (0.10) \\
\cline{2-7}
& P-DQN $R^{prob}$ & 0.81 (0.33) & 2.48 (0.47) & 1.57 (0.71) & 0.56 (0.09) & \textbf{0.74} (0.14) \\
\hline
\end{tabular}
}
\end{table}

\section{Conclusion and Future Work}
\label{sec:conclusion}
In this study, we focus on the problem of learning a policy to generate Sequential Counterfactuals (SCF) using Reinforcement Learning (RL). We identify issues in related methods that lead to policies over-utilizing features, thus becoming ineffective in practice by repeatedly taking identical actions for any input. To address this issue, we propose taking continuous actions and implementing a denser reward function that considers the classifier's output probabilities instead of relying only on class labels. By introducing an action entropy metric, we quantify the extent of feature over-utilization and demonstrate the effectiveness of our approach in mitigating this problem. These findings highlight the importance of carefully designing the problem formulation, reward function, and modeling choices, particularly when generating SCFs. 

For future research, we will focus on generating SCFs for mixed features without transforming them into a common data type. We aim to construct RL agents that work in mixed action spaces and explore distance metrics that take mixed feature interdependencies into account.

\section{Acknowledgements}
Our work is Funded by the Deutsche Forschungsgemeinsschaft (DFG, German Research Foundation ) - SFB1463 - 434502799. I further acknowledge the support by the European Union, Horizon Europe project MAMMOth under contract number 101070285.

%
%
%
\bibliographystyle{splncs04}
\bibliography{bibliography}
%
\end{document}